\documentclass[runningheads]{llncs}
\usepackage[backend=biber,style=numeric,sorting=none]{biblatex}
\usepackage[acronym,toc]{glossaries}
\newacronym{ai}{AI}{Artificial Intelligence}
\newacronym{yolo}{YOLO}{You Only Look Once}
\newacronym{cnn}{CNN}{Convolutional Neural Network}
\newacronym{dl}{DL}{Deep Learning}
\newacronym{sota}{SotA}{State-of-the-Art}
\newacronym{uav}{UAV}{Unmanned Aerial Vehicle}
\newacronym{iwm}{IWM}{Integrated weed Management}
\newacronym{sswm}{SSWM}{Site-Specific Weed Management}
\newacronym{njn}{NJN}{NVIDIA Jetson Nano}
\newacronym{rtagcm}{RT-ABGCM}{Real-Rime Autonomous Black-Grass Classification and Mapping}
\newacronym{cuda}{CUDA}{Compute Unified Device Architecture}
\newacronym{gmm}{GMM}{Gaussian Mixture Model}
\newacronym{apu}{APU}{Application Processor Unit}
\newacronym{cpu}{CPU}{Central Processor Unit}
\newacronym{gpu}{GPU}{Graphics Processing Unit}
\newacronym{rgbd}{RGBd}{Red, Green, Blue and depth}
\newacronym{wnb}{WnB}{Weights and Biases}
\newacronym{map}{mAP}{mean Average Precision}
\newacronym{tp}{TP}{True Positive}
\newacronym{tn}{TN}{True Negative}
\newacronym{fp}{FP}{False Positive}
\newacronym{fn}{FN}{False Negative}
\newacronym{fps}{FPS}{Frames Per Second}
\newacronym{ros2}{ROS2}{Robotic Operating System 2}
\usepackage{url}
\usepackage{xcolor}
\usepackage{hyperref}
\hypersetup{
 colorlinks,
 linkcolor={red!50!black},
 citecolor={blue!50!black},
 urlcolor={blue!80!black}
}
\usepackage{soul}
\usepackage{lineno}
\modulolinenumbers[1]
\usepackage{float}
\usepackage{subcaption}
\usepackage{caption}

\usepackage[T1]{fontenc}
\usepackage{graphicx}
\begin{document}
\title{WeedScout: Real-Time Autonomous blackgrass Classification and Mapping using dedicated hardware}
%
%
\author{Matthew Gazzard\inst{1} \and
Helen Hicks \inst{2}\orcidID{0000-0003-1325-2293} \and
Isibor Kennedy Ihianle\inst{1}\orcidID{0000-0001-7445-8573} \and
Jordan J. Bird\inst{1}\orcidID{0000-0002-9858-1231} \and
Md Mahmudul Hasan\inst{3}\orcidID{0000-0003-2543-3112} \and
Pedro Machado\inst{1}\orcidID{0000-0003-1760-3871}}
\authorrunning{M. Gazzard et al.}
%
\institute{Computational Intelligence Applications Research Group, Nottingham Trent University, Clifton Campus, NG11 8NS, UK, \url{https://www.ntu.ac.uk/research/groups-and-centres/groups/computational-intelligence-applications-research-group}\\ \email{n0932346@my.ntu.ac.uk,\{isibor.ihianle,jordan.bird,pedro.machado\}@ntu.ac.uk} \and
School of Animal, Rural and Environmental Sciences, Nottingham Trent University, Brackenhurst, Southwell, NG25 0QF, UK
\\
\url{https://www.ntu.ac.uk/study-and-courses/academic-schools/animal,-rural-and-environmental-sciences} \\
\email{helen.hicks@ntu.ac.uk} \and 
Mediprospects AI, 5-7 High Street, London, E13 0AD, UK\\
\url{https://www.mediprospects.ai/} \\
\email{m.mahmudul@mediprospects.ai}
}
\maketitle 
\begin{abstract}
Blackgrass (Alopecurus myosuroides) is a competitive weed that has wide-ranging impacts on food security by reducing crop yields and increasing cultivation costs. In addition to the financial burden on agriculture, the application of herbicides as a preventive to blackgrass can negatively affect access to clean water and sanitation. The WeedScout project introduces a \gls*{rtagcm}, a cutting-edge solution tailored for real-time detection of blackgrass, for precision weed management practices. Leveraging \gls*{ai} algorithms, the system processes live image feeds, infers blackgrass density, and covers two stages of maturation. The research investigates the deployment of \gls*{yolo} models, specifically the streamlined \gls*{yolo}v8 and \gls*{yolo}-NAS, accelerated at the edge with the \gls*{njn}. By optimising inference speed and model performance, the project advances the integration of \gls*{ai} into agricultural practices, offering potential solutions to challenges such as herbicide resistance and environmental impact. Additionally, two datasets and model weights are made available to the research community, facilitating further advancements in weed detection and precision farming technologies.

\end{abstract}
%
%
%

\section{Introduction}\label{s1:introduction}
Blackgrass (Alopecurus myosuroides) is a problematic weed that has potential impact on food security due to its interruption of oilseed rape and several winter cereals. As weed species became increasingly resistant to herbicides, managing weed populations became more difficult. Varah et al. \cite{varah2020} concluded that herbicide resistant blackgrass resulted in £400 million lost gross profit and 0.8 million tonnes lost yield in England alone. Furthermore, Varah et al. \cite{varah2020} predicted that in the absence of herbicide control, losses could reach £1 billion and 3.4 million tonnes annually.
In this paper, the WeedScout project presents an \acrfull*{ai} solution tailored to detect blackgrass in wheat crops, streamline farmers' identification processes, and resource allocation. The innovative approach not only facilitates precision weed management, but also holds promise in reducing herbicide costs and hindering the advancement of herbicide resistance. The contributions of the paper include the compilation and sharing of a diverse blackgrass dataset, employing state-of-the-art \gls*{dl} algorithms, and the deployment of the models on dedicated hardware. Initially, our focus was on exploring the use of \gls*{ai} in object detection, with particular emphasis on real-time implementations, given the challenge of accurately mapping blackgrass. Furthermore, our work contributes annotated two blackgrass datasets \cite{gazzard2024a,gazzard2024b} to support future work. The WeedScout delivers \acrfull*{rtagcm} powered by the \acrfull*{njn}. By leveraging the computational prowess of the \gls*{njn}, \gls*{ai} inference is accelerated at the edge, enabling swift detection and analysis of blackgrass. Moreover, the proposed approach not only maps the blackgrass density but also provides real-time geolocation, obviating the necessity for a continuous internet connection. Thus, WeedScout offers a comprehensive and efficient means of managing blackgrass infestations in wheat fields.

The paper is organised into the following sections: section \ref{s2:lr} reviews relevant works; section \ref{s3:methodology} elaborates on the proposed methodology; section \ref{s4:results} conducts an analysis of the results; and, section \ref{s5:conclusions} presents the concluding remarks and outlines future research directions.

\section{Literature Review}\label{s2:lr}
Weeds are commonly recognised as the most significant biotic factor affecting crop production \cite{scavo2020}. However, it is noteworthy that low densities of weeds can also provide agronomic and ecological benefits \cite{scavo2020}. Therefore, weeds should be viewed as agricultural entities that require management rather than outright elimination. Historically, herbicides have been the most widely employed method for weed management, aimed at maximising crop yield while minimising human effort and cost \cite{varah2020,scavo2020}. However, extensive use of herbicides has led to the evolution of resistance on a large scale, jeopardising yield benefits as herbicide efficacy decreases \cite{varah2020,scavo2020,hicks2018,comont2020}. Resistance to herbicides presents a significant problem, as resistance to one class of herbicides can indicate resistance to multiple types, and the use of herbicide mixtures in hopes of reducing specialist resistance can paradoxically lead to more general resistance \cite{hicks2018,comont2020}. Such reliance on herbicides as the primary weed management strategy is unsustainable, given the expense and time required to formulate new weed-killing chemicals, coupled with the inevitable development of resistance rendering them obsolete \cite{hicks2018}. Additionally, it is crucial to acknowledge the adverse environmental impacts of herbicides, which harm biodiversity, water, and soil quality \cite{varah2020}.

The abundance of A. myosuroides weeds is currently mediated by herbicide resistance \cite{hicks2018}, a phenomenon supported by Varah et al. \cite{varah2020}. The consequential impact of A. myosuroides on yield stands out as the most economically significant weed in Western Europe due to its high yield impact \cite{hicks2018}. Comont et al. \cite{comont2020} discuss how glyphosate has emerged as one of the final lines of defence in controlling A. myosuroides, yet A. myosuroides is already developing resistance to the herbicide. Resistance to herbicides mirrors the broader trend of resistance in agricultural sectors, posing a substantial challenge. Given the increasing difficulties in herbicide-based weed management, it becomes imperative to explore more diverse methods of weed management that integrate targeted site management, an area where \gls*{ai} proves invaluable.

Ahmed et al. \cite{ahmed2021} delineate the concept of challenging environments which includes occlusion, poor lighting, and objects merging into backgrounds as typical scenarios where object detection using \gls*{ai} has historically struggled. Leveraging the definition alongside Hasan et al. \cite{hasan2021}, crop fields emerge as exemplary instances of such environments. Although two-stage detectors generally offer higher accuracy, their slower processing speed renders them unsuitable for real-time applications \cite{zaidi2022}. Given the planned integration of embedded or edge devices in the proposed blackgrass detection solution, it becomes imperative to employ object detection methodologies optimised for limited computational resources \cite{ahmed2021}. Consequently, the utilisation of two-stage detectors is discounted, a notion corroborated by Zaidi et al. \cite{zaidi2022}, who emphasise the importance of lightweight models. When considering object detection algorithms, \acrfull*{yolo} is lauded for its adeptness in striking a balance between speed and accuracy, rendering it ideal for real-time object detection \cite{terven2023,pomykala2022,brandenburg2020}. While other models may offer superior accuracy, the equilibrium holds significant value. Moreover, \gls*{yolo} provides a range of model scales to tailor its performance to specific application requirements and hardware capabilities \cite{terven2023,pomykala2022,brandenburg2020}. Therefore, \gls*{yolo} models have already found deployment on edge devices such as robots and drones and boast a history of successful implementation in agricultural contexts \cite{terven2023,,pomykala2022,brandenburg2020}.

In agriculture, \gls*{ai} deployment for weed detection and mapping is increasingly prevalent, as evidenced by studies such as those conducted by Krähmer et al. \cite{krahmer2020}, Xu et al. \cite{xu2020}, Jabir and Falih \cite{jabir2022}, Hasan et al. \cite{hasan2021}, Wu et al. \cite{wu2021}, Xu et al. \cite{xu2023}, and Dainelli et al. \cite{dainelli2023}. Krähmer et al. \cite{krahmer2020} further specify that herbicide resistance distribution can also be effectively mapped. In the context of wheat crops, Xu et al. \cite{xu2020} elaborate on the tillering stage as the opportune time for weed detection due to reduced occlusion between wheat and weeds, a factor alleviating some challenges encountered in challenging environments \cite{ahmed2021,hasan2021}. However, despite the availability of numerous datasets, challenges persist in training models specifically for weed detection within crop fields. Dainelli et al. \cite{dainelli2023} highlight limitations in data collection methods and availability, stressing the costliness and time-consuming nature of creating custom datasets. Krähmer et al. \cite{krahmer2020} underscore the difficulty in identifying growth stages and the challenge detection algorithms face in distinguishing between similar weeds and a plethora of unknown species \cite{kavhiza2020}. Nonetheless, finding an automated solution for weed detection remains imperative, as manual methods prove economically unviable \cite{kavhiza2020}. While the hardware for weed monitoring may incur initial expenses, the resulting reductions in herbicide usage offset costs and offer environmental benefits. Moreover, the inefficiency of manual weed detection underscores the potential of technologies such as \glspl*{uav} or robots to supplement or replace manual labor. \glspl*{uav} equipped with high-resolution cameras are particularly effective in detecting low-density weed patches \cite{wu2021}. However, limitations exist, as most \gls*{uav} applications involve capturing images solely above the crop canopy, potentially missing crucial information \cite{krahmer2020}. Thus, a solution closely monitoring wheat fields is desirable. Examples of such solutions already exist, contributing to \gls*{iwm} and \gls*{sswm}. For instance, BonnBot-I \cite{ahmadi2022} serves as a precision weed management and field monitoring platform, utilising ground-based rovers for weed detection and herbicide application. Moreover, Mattivi et al. \cite{mattivi2021} exemplify the successful deployment of cost-effective \glspl*{uav} equipped with open-source software for implementing smart farming techniques, even on small-scale farms. These systems play pivotal roles in advancing Agriculture 4.0 initiatives \cite{wu2021,xu2023}. Localised herbicide application can be realised through the utilisation of weed maps or real-time sensors \cite{kavhiza2020}. The WeedScout project extends upon the literature reviewed in the present section by perform \gls*{rtagcm}, independent of internet connectivity, and leveraging edge acceleration with the \gls*{njn}.

\section{Methodology}\label{s3:methodology}
The WeedScout system receives live image feeds, executes \gls*{ai} inference on the \gls*{njn}, and produces an output depicting the density of detected blackgrass, potentially discerning various stages of maturation. The chapter provides a detailed description of the \gls*{rtagcm} module which is the core of the WeedScout.

As discussed in the previous section, \gls*{yolo} models are well-known for their efficacy in weed detection. However, due to limited computational resources on edge devices, larger \gls*{ai} models, despite their higher accuracy compared to smaller ones, are not feasible for deployment. Moreover, larger \gls*{yolo} models exhibit increased latency, rendering them unsuitable for real-time detection systems. To determine the veracity of the effectiveness of \gls*{yolo}-NAS, \gls*{yolo}v8 models were also trained on the same two custom datasets \cite{gazzard2024a,gazzard2024b} and benchmarked for determining which model type works best in the problem domain.

The developed models were deployed on the cost-effective \gls*{njn}. The compact computer is purpose-built for \gls*{ai} applications and leverages \gls*{cuda}\footnote{Available online, \protect\url{https://developer.nvidia.com/cuda-zone}, last accessed: 08/05/2024} for accelerating AI operations. However, given the edge deployment's limited resources, training the model directly on the device is impractical. Therefore, the models will undergo training on a separate, more powerful machine. The models performed inference on video inputs of blackgrass, generating video output with overlays of the model's detections.

\vspace*{-5mm} 
\begin{figure}[]
	\centering
	\includegraphics[scale=0.8]{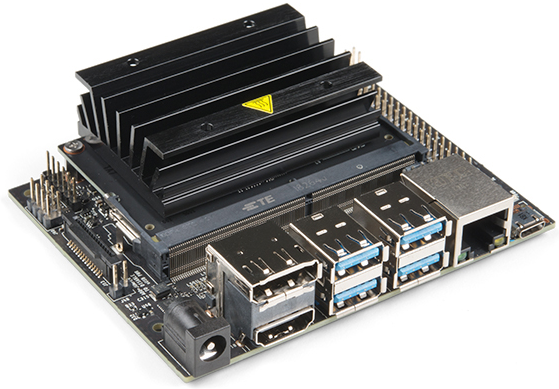}
	\caption{NVIDIA Jetson Nano.}
 \label{fig:njn}
\end{figure}
\begin{figure}[]
\vspace*{-5mm}
    \centering
    \begin{tabular}{cc}
        \begin{minipage}{0.45\textwidth}
            \centering
            \includegraphics[scale=0.10]{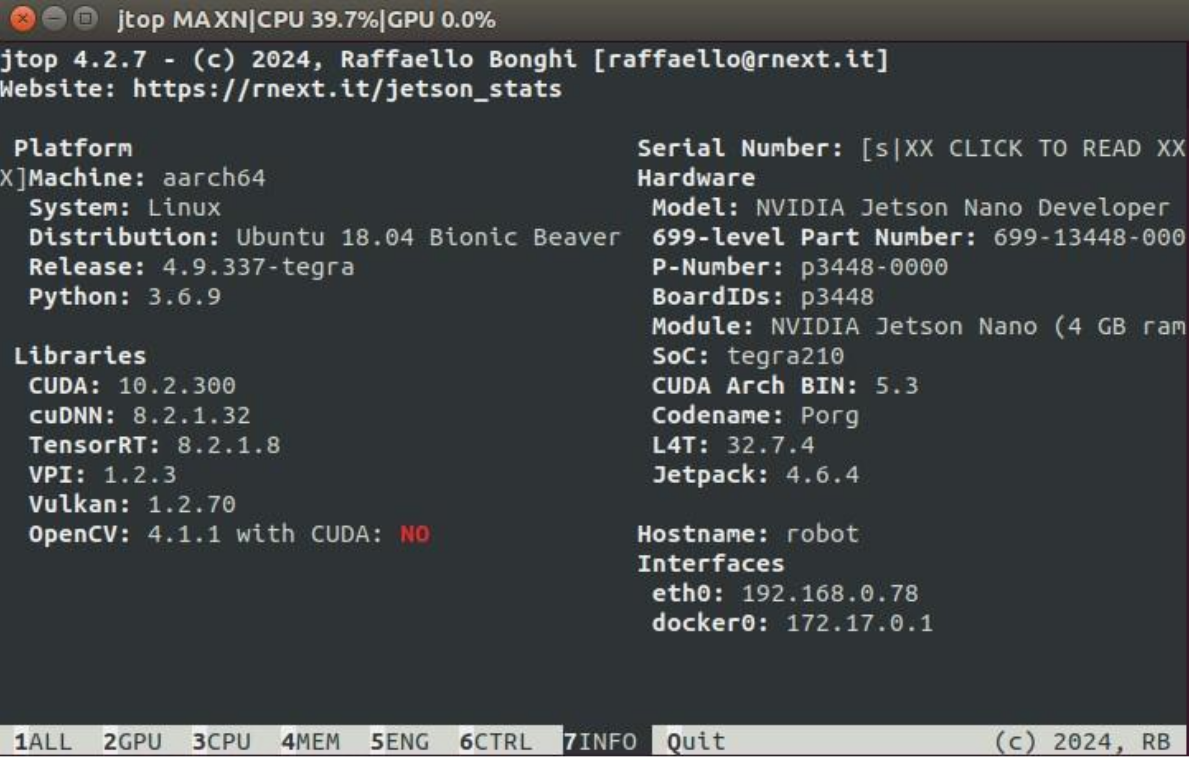} \caption{NVIDIA JetPack SDK 2.0.}
 \label{fig:njn_jetpack}
        \end{minipage}
        &
        \begin{minipage}{0.45\textwidth}
            \centering
            \includegraphics[scale=0.10]{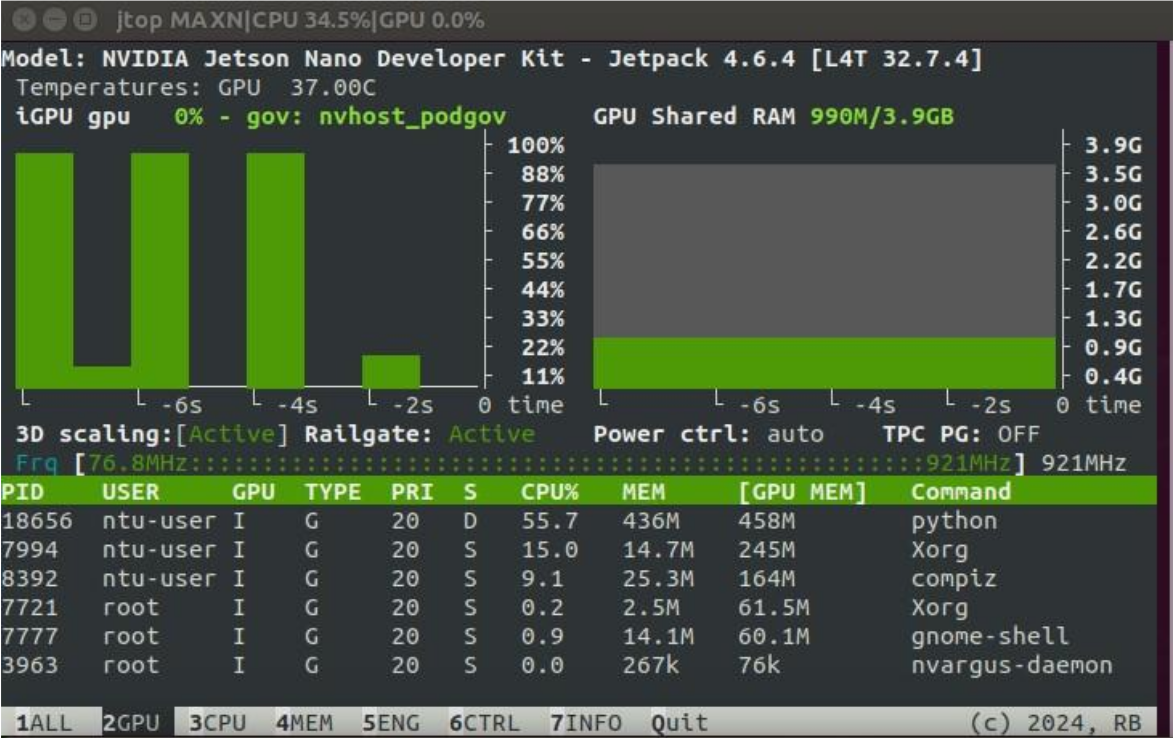}
            \caption{\gls*{njn} AI inference.}\label{fig:njn_inference}
        \end{minipage} \\
    \end{tabular}
\end{figure}

\vspace*{-5mm} 
The \gls*{njn} (see Fig. \ref{fig:njn}) was flashed with JetPack SDK 2\footnote{Available online, \protect/url{https://developer.nvidia.com/embedded/jetpack}, last accessed: 08/05/2024} (see Fig. \ref{fig:njn_jetpack}), running \gls*{ros2} Dashing\footnote{Available online, \protect\url{https://docs.ros.org/en/dashing/index.html}, last accessed 08/05/2024} powered by Ubuntu 18.04.06 LTS \footnote{Available online, \protect\url{https://releases.ubuntu.com/18.04/}, last accessed: 08/05/2024} (see Figure 6). The JetPack SDK provides all the NVIDIA libraries necessary for \gls*{ai} edge applications and \gls*{ros2} is necessary for compatibility with mobile robotic platforms.

\vspace*{-10mm} 
\subsection{WeedScout Platform Workflow}
\begin{figure}[]
\vspace*{-5mm}
	\centering
	\includegraphics[scale=0.2]{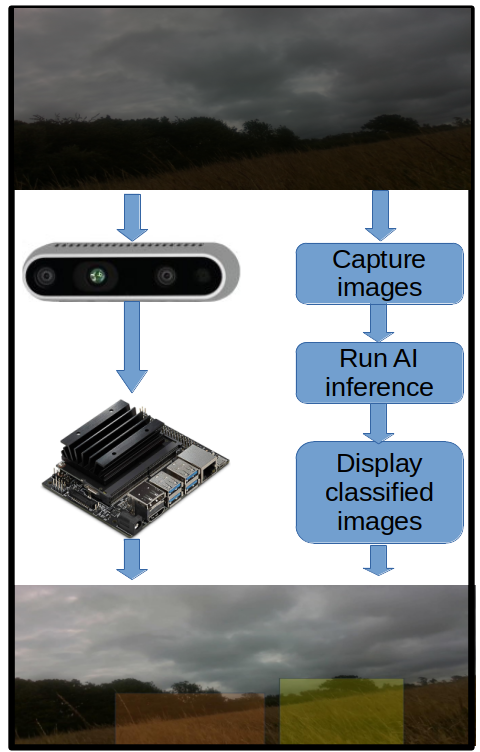}
	\caption{The proposed WeedScout architecture workflow.}
 \label{fig:weedscout}
\end{figure}

A meticulously designed and implemented prototype for \gls*{rtagcm} is depicted in Figure \ref{fig:weedscout}. The system integrates an \gls*{njn} equipped with an Intel RealSense D435i depth sensor. The \gls*{njn} serves as the computational backbone for accelerating the \gls*{yolo}v8 and \gls*{yolo}-NAS inference at the edge. Specifically tailored for a myriad of tasks, the \gls*{njn} handles image identification, object detection, data segmentation, and speech processing. It boasts a robust configuration, featuring a 128-core Maxwell \gls*{gpu}, a quad-core ARM A57 \gls*{apu} operating at 1.43 GHz, and 4 GB of 64-bit LPDDR4 memory running at 25.6 GB/s. Moreover, the system is bolstered by the inclusion of a pre-trained \gls*{dl} models, enhancing its detection capabilities. 
\vspace*{-5mm}
While semantic segmentation would be beneficial for clearly delineating boundaries between blackgrass and wheat, annotating a sufficiently large dataset would be exceedingly time-consuming due to the likelihood of images containing numerous individual plants. Annotating images for object detection is comparatively simpler as it does not require drawing complex polygons around the desired objects. Moreover, it was determined that employing object detection would not adversely affect the performance of the generated models. Therefore, object detection was chosen as the computer vision methodology.

Object detection necessitates annotating a given dataset by drawing bounding boxes around objects within images and assigning each box to the corresponding object class. Roboflow\footnote{Available online, \protect\url{https://roboflow.com/}, last accessed: 10/05/2024} was used to increase productivity. Leveraging Roboflow proved advantageous as it facilitated the easy management and storage of datasets; annotated images could be categorised into training, validation, and test sets before undergoing augmentation and export.

The dataset for mature blackgrass was collected using an Intel RealSense D435i\footnote{Available online, \protect\url{https://www.intelrealsense.com/depth-camera-d435i/}, last accessed: 10/05/2024} at a wheat field in England, UK. The Intel RealSense D435i was used to capture \gls*{rgbd} data and stored in a bag file. Consequently, it was necessary to extract standard video footage from the file using Intel RealSense SDK\footnote{Available online, \protect\url{https://www.intelrealsense.com/sdk-2/}, last accessed: 10/05/2024}. The process yielded thousands of PNGs of standard images alongside thousands of images containing depth data. Since the depth data was unnecessary for the proposed work, it was discarded, and a random selection of 780 PNGs was extracted using a PowerShell script to form an initial dataset.

Following the loading of the 780 images into Roboflow, 156 (20\%) were allocated for validation, and 78 (10\%) for testing, leaving the remaining 546 (70\%) for training. Subsequently, all images were annotated using Roboflow’s annotation tool, categorising blackgrass into three different density categories: low, medium, and high. To augment the size and diversity of the initial dataset, data augmentation techniques were applied to decrease the likelihood of overfitting and enhance model performance. Augmentations were applied to both the images and the bounding box annotations (refer to Table \ref{tab:augmentations}). Additionally, a version of the dataset without augmentations was preserved for training, enabling performance comparison. 
\begin{table}[]
\vspace*{-10mm} 
\centering
\caption{Dataset augmentations}\label{tab:augmentations}
\begin{tabular}{|c|c|c|}
\hline\hline
Type & Applied to & Option \\ \hline
Flip & Image, Bounding Box & Horizontal, Vertical \\ \hline
90° Rotate & Image, Bounding Box & Clockwise, Counter-Clockwise, Upside, Down \\ \hline
Crop & Image, Bounding Box & 0\% Minimum Zoom, 20\% Maximum Zoom \\ \hline
Rotation & Image, Bounding Box & Between -15° and +15° \\ \hline
Shear & Image, Bounding Box & ±15° Horizontal, ±15° Vertical \\ \hline
Hue & Image & Between -25° and +25° \\ \hline
Saturation & Image & Between -30\% and +30\% \\ \hline
Brightness & Image, Bounding Box & Between -25\% and +25\% \\ \hline
Exposure & Image, Bounding Box & Between -14\% and +14\% \\ \hline
Blur & Image & Up to 2.5px \\ \hline
Noise & Image, Bounding Box & Up to 1.99\% of pixels \\ \hline
Cutouth & Image & 3 boxes with 10\% size  \\ \hline
Mosaic & Image & Applied \\ \hline
\end{tabular}
\end{table}
\vspace*{-5mm}
A separate approach was also explored on Roboflow utilising blackgrass seedling dataset obtained from the Plant seedling datasets available on Kaggle \footnote{Available online, \protect\url{https://www.kaggle.com/datasets/vbookshelf/v2-plant-seedlings-dataset}, last accessed: 10/05/2024}. The images of the plant seedling dataset were annotated, and data augmentation was applied due to the limited size of the dataset (see Table \ref{tab:augmentations}). The seedling dataset comprised only one class, blackgrass as it was not intended for detecting weed density. Instead, it was envisaged that any resulting model could be used for precision weed management in newly planted wheat fields.

Developing \gls{yolo}-NAS models required the utilisation of the open-source model library SuperGradients\footnote{Available online, \protect\url{https://github.com/Deci-AI/super-gradients}, last accessed: 10/05/2024}. In addition to facilitating the training of regular models, the library enables the SuperGradients libraries to quantitise models through its post-training quantisation and quantization aware training functionalities. SuperGradients also provides the ability to connect third-party monitoring software to display the metrics tracked during model training and validation. Given the iterative nature of training \gls*{ai} models, where adjustments to the dataset or hyperparameters may be necessary based on training outcomes, it is crucial to monitor performance at both training and validation stages to assess the impact of any modifications. \gls*{wnb} serves as a platform for storing \gls*{ai} model data, facilitating the comparison and evaluation of model performance. Any monitoring data collected during model training was hosted on \gls*{wnb}, enabling the automatic generation of graphs for essential metrics such as \gls*{map}, precision, and recall for use in model comparison. The use of \gls*{wnb} also served as the project's test suite, as the performance metrics of the models are intrinsically linked to the successful video inference.
Although training AI models is resource intensive, maximising the use of the hardware used for training is crucial to ensuring models are trained within an acceptable timeframe. Although it is possible to train \gls*{ai} models using a computer's \gls*{cpu}, the method is not the most efficient. Leveraging \gls*{cuda} with an NVIDIA \gls*{gpu} can significantly reduce training time by accelerating processing through \gls*{gpu}-acceleration. Thus, training with \gls*{gpu}-acceleration is the preferred methodology, either by using available resources or using cloud solutions such as Google Collab.

Using a \gls*{gpu} offers benefits beyond accelerated training; it also enables faster model inference. This is essential as any recorded footage of blackgrass should be inferred at the fastest possible speed to maintain usability. The chosen training machine utilised an NVIDIA GeForce RTX 3080 10GB \gls*{gpu} with \gls*{cuda} version 12.4 (see Figure 13). Before starting training, it was necessary to prepare the training machine by downloading the \gls*{cuda} Toolkit, as it does not come pre-installed. SuperGradients utilises PyTorch, a Python-based machine learning framework. To enable PyTorch to utilise \gls*{cuda}, it is necessary to download versions of it that come packaged explicitly with \gls*{cuda} capabilities. Without these specific packages, training and inference using NVIDIA \glspl*{gpu} would be impossible.

\begin{figure}[]
\vspace*{-5mm}
	\centering
	\includegraphics[scale=0.1]{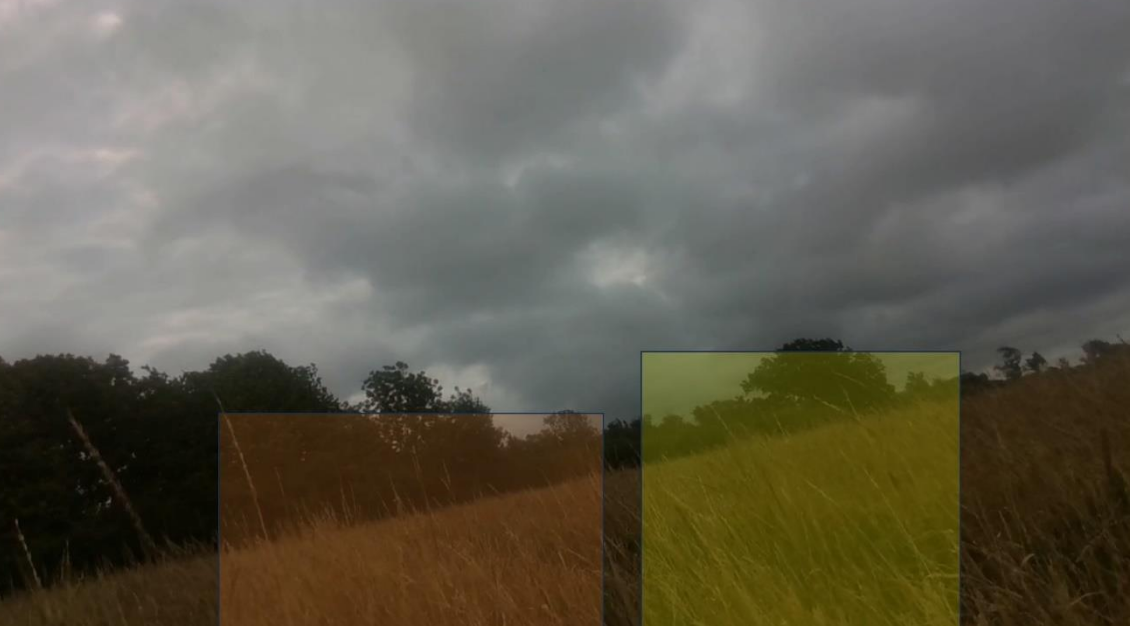}
	\caption{Blackgrass detection overlay initial design. Yellow is being used to demonstrate low density and orange is being used to demonstrate medium density. High density patches would have a red overlay.}\vspace*{-5mm}
 \label{fig:overlay}
\end{figure}

Before deploying models to the \gls*{njn}, it was essential to devise a design for presenting the blackgrass detections to users. By default, the prediction function of SuperGradients draws labelled bounding boxes around the detections, indicating the confidence threshold of the prediction (see Figure \ref{fig:overlay}). However, the objective of blackgrass inference is to identify areas with varying densities of blackgrass weeds, and SuperGradients does not offer a density overlay option. Additionally, SuperGradients cannot be used to infer YOLOv8 models, necessitating the creation of a custom overlay.
Inspiration was drawn from heatmaps, where higher densities are represented by increasingly warmer colours. For instance, lower densities may be highlighted in yellow, while higher densities can be depicted in orange. Areas with the highest densities could have a red overlay to denote their severity. Figure \ref{fig:overlay} was devised to provide guidance for designing the output of model inference returned to users.

\section{Results Analysis}\label{s4:results}
In this section, we will discuss the diverse results obtained for the two datasets tested against both \gls*{yolo}v8 and \gls*{yolo}-NAS.

\subsection{blackgrass Seedling dataset}
To evaluate the suitability of the seedlings dataset (see Table \ref{tab:seed_yolo}) tested against the \gls*{yolo}v8.

\begin{table} []
\vspace*{-10mm}
\caption{Number of images within augmented blackgrass seedling datasets}\label{tab:seed_yolo}
\centering
\begin{tabular}{|c|c|c|c|}
\hline
Train & Validation & Test & Total \\\hline
2820 & 168 & 112 & 3100 \\\hline
\end{tabular}
\vspace*{-5mm}
\end{table}

For the seedling dataset, there was only the blackgrass class to be detected. From the results obtained, the \gls*{tp} value of 0.81 indicated the model's success, as the vast majority of blackgrass in the positive samples was detected. While metrics for \gls*{map} quickly converged within approximately 20 epochs, the values failed to reach 1.0 The validation loss (\ref{fig:cls_seed_yolo}) gradually decayed as expected, but training loss exhibited greater stability than its validation counterparts. Nevertheless, the results from creating a \gls*{yolo}-NAS model using the blackgrass seedling dataset provided clearer insights into overfitting compared to the \gls*{yolo}v8 model. Convergence was remarkably swift for both \gls*{map} and recall during training and validation. The final values for \gls*{map} and recall metrics during training were also strikingly close to 1. While validation precision remained high, it was noticeably lower compared to its training counterpart, which tended towards 1.0 like training \gls*{map} and recall. Therefore, an interesting trend was observed when comparing the training loss and validation loss, where training loss decreased while validation loss increased. The results indicate that the model was overfitting. Combined with other trends observed during both \gls*{yolo}-NAS and \gls*{yolo}v8 training and validation, it was concluded that overfitting negatively impacted the performance of models trained on the blackgrass seedling dataset. Unfortunately, beyond the existing data augmentations, there was no effective way to further increase the size and diversity of the blackgrass seedling dataset. Therefore, the focus shifted to prioritising the detection of mature blackgrass.

\subsection{Mature blackgrass dataset}
The mature blackgrass dataset (see Table \ref{tab:dataset2} was generated using the same augmentations described in Table \ref{tab:augmentations}.
\begin{table}[]
\vspace*{-10mm}
\centering
\caption{Dataset Details}
\label{tab:dataset2}
\begin{tabular}{|c|c|c|c|c|}
\hline\hline
Dataset & Train & Validation & Test & Total \\ \hline
Blackgrass mature & 6552 & 624 & 312 & 7488 \\ \hline
\end{tabular}
\vspace*{-5mm}
\end{table}

Given the state-of-the-art performance typically associated with \gls*{yolo}-NAS models, it was expected that similar models would surpass \gls*{yolo}v8 in later epochs. However, the \gls*{map} for the \gls*{yolo}v8 model demonstrated a slight improvement compared to its \gls*{yolo}-NAS counterpart. Furthermore, the recall for the \gls*{yolo}v8 model outperformed \gls*{yolo}-NAS when running on NVIDIA Jetson nano, although both precision and recall to be similar. The observation is intriguing considering that the \gls*{yolo}v8 nano model. The discrepancy in size makes it obvious that the \gls*{yolo}v8 model has a better performance than the \gls*{yolo}-NA model. Additionally, the absence of loss escalation during validation with the \gls*{yolo}v8 model contrasts with the trend observed in the \gls*{yolo}-NAS models, where validation loss increases.
 
The models contained in Table \ref{tab:infer_conf} ran inference on a video containing high densities of blackgrass.
\begin{table}[]
\centering
\caption{Models used in inference testing and the confidence used to perform inference.}\label{tab:infer_conf}
\begin{tabular}{|l|c|c|}
\hline
Model & Confidence & \begin{tabular}[c]{@{}c@{}}Frames with\\ detections\end{tabular} \\ \hline
\gls*{yolo}-NAS & 0.7 & 444 \\ \hline
\gls*{yolo}v8 & 0.7 & 799 \\ \hline
\end{tabular}
\vspace*{-5mm}
\end{table}
To assess the run-time performance of the mature blackgrass models, a brief 5-second video was employed for inference. Upon examining the inference times of various models listed in Table \ref{tab:infer}, it became evident that the \gls*{yolo}v8 model, trained on the mature blackgrass dataset, exhibited notably faster inference speeds compared to the \gls*{yolo}-NAS models. The discrepancy can be attributed to the variance in model size and possibly to differences in the efficiency of prediction functions between Ultralytics\footnote{Available onine, \protect\url{https://docs.ultralytics.com/}, last accessed: 12/05/2024} and SuperGradients.

In order to utilise \gls*{gpu}, the \gls*{njn} was configured to run the inference using the pre-trained \gls*{yolo}v8 weights. The runtime for inference on the same video used for training on the testing machine is presented in Table \ref{tab:infer}). Although the inference time is notably longer, the results were anticipated as \gls*{yolo}-NAS runs on the \gls*{cpu} while \gls*{yolo}v8 utilises \gls*{gpu}. Given the significantly longer inference time compared to the supplied source video, optimisations to the model and code are imperative to reduce inference time and potentially enable real-time detection.

\begin{table}[]
\vspace*{-5mm}
\centering
\caption{Mean average time for mature blackgrass models to inference a 5s long video on the \gls*{cpu} only, and \gls*{cpu} and \gls*{gpu} accelerated.}\label{tab:infer}
\begin{tabular}{|l|c|c|c|c|}
\hline
Model & \begin{tabular}[c]{@{}c@{}}Inference 1 \\ {[}s{]}\end{tabular} & \begin{tabular}[c]{@{}c@{}}Inference 2\\ {[}s{]}\end{tabular} & \begin{tabular}[c]{@{}c@{}}Inference 3\\ {[}s{]}\end{tabular} & \begin{tabular}[c]{@{}c@{}}Mean inference\\ {[}s{]}\end{tabular} \\ \hline
\begin{tabular}[c]{@{}l@{}}\gls*{yolo}v8 (\gls*{cpu} and \gls*{gpu})\end{tabular} & 82.98 & 86.25 & 82.69 & 83.97 \\ \hline
\begin{tabular}[c]{@{}l@{}}\gls*{yolo}-NAS (\gls*{cpu} only)\end{tabular} & 229.71 & 223.75 & 217.93 & 223.80 \\ \hline
\end{tabular}
\vspace*{-5mm}
\end{table}
\section{Conclusions and Future Work}\label{s5:conclusions}
This paper introduces a novel \gls*{rtagcm} designed to operate effectively at the edge. Following multiple iterations of dataset augmentation and annotation, two types of models were developed capable of successfully detecting blackgrass. Given constraints with the blackgrass seedling dataset, the emphasis shifted towards detecting mature blackgrass. Surprisingly, YOLOv8 proved more effective for the task compared to \gls*{yolo}-NAS. However, both models encountered challenges in detecting low and medium density blackgrass.The challenge arose from the mature blackgrass dataset's significant bias towards high-density patches, making it challenging to generate annotated datasets containing an adequate number of images depicting lower densities for robust training. Rectifying the challenge requires obtaining a more diverse dataset and conducting additional experimentation to identify the best mix of training hyperparameters and dataset augmentations for refining density detection models. Despite these challenges, models were successfully deployed on the \gls*{njn}, demonstrating the efficacy of the model production pipeline. While the \gls*{njn} enables inference, it operates at sub-optimal \gls*{gpu} acceleration, resulting in slower inference times that preclude real-time performance.

Inference speed of the current solution could be improved by investigating other methods of model inferencing. Other lightwight \gls*{ai} frameworks such as TensorRT and DeepStream SDK could be utilised to improve the frame rate. Improving the \gls*{fps} is necessary for real-time inference. If real-time inference at a stable \gls*{fps} was achieved, it would be possible to create an automated sprayer or electric discharge combined with more research into detecting blackgrass seedlings so that blackgrass could be tackled at the earliest possible stage.
\begin{credits}
\subsubsection{\discintname}
The authors have no competing interests to declare that they are
relevant to the content of this article.
\vspace*{-5mm}
\end{credits}
%
%
%
%
\addcontentsline{toc}{section}{References}
\printbibliography
\end{document}